\documentclass{article}

\usepackage{PRIMEarxiv}

\usepackage[utf8]{inputenc} 
\usepackage[T1]{fontenc}    
\usepackage{hyperref}       
\usepackage{url}            
\usepackage{booktabs}       
\usepackage{amsfonts}       
\usepackage{nicefrac}       
\usepackage{microtype}      
\usepackage{lipsum}
\usepackage{fancyhdr}       
\usepackage{amsmath}
\usepackage{graphicx}       
\graphicspath{{media/}}     
\usepackage{algorithm}
\usepackage{algorithmic}
\usepackage{float}
\usepackage{subfig}
\usepackage[toc, page]{appendix}
\usepackage{multirow}
\usepackage{longtable}
\graphicspath{ {./images/} }

\title{Sampler Scheduler for Diffusion Models
}

\author{
  Zitong Cheng \\
  Sun Yat-sen University \\
  \texttt{chengzt5@mail2.sysu.edu.cn} \\
}

\begin{document}
\maketitle

\begin{abstract}
Diffusion modeling (DM) has high-quality generative performance, and the sampling problem is an important part of the DM performance. Thanks to efficient differential equation solvers, the sampling speed can be reduced while higher sampling quality is guaranteed. However, currently, there is a contradiction in samplers for diffusion-based generative models: the mainstream sampler choices are diverse, each with its own characteristics in terms of performance. However, only a single sampler algorithm can be specified on all sampling steps in the generative process. This often makes one torn between sampler choices; in other words, it makes it difficult to fully utilize the advantages of each sampler. In this paper, we propose the feasibility of using different samplers (ODE/SDE) on different sampling steps of the same sampling process based on analyzing and generalizing the updating formulas of each mainstream sampler, and experimentally demonstrate that such a multi-sampler scheduling improves the sampling results to some extent. In particular, we also verify that the combination of using SDE in the early sampling steps and ODE in the later sampling steps solves the inherent problems previously caused by using both singly. We show that our design changes improve the sampling efficiency and quality in previous work. For instance, when Number of Function Evaluations (NFE) = 24, the ODE Sampler Scheduler achieves a FID score of 1.91 on the CIFAR-10 dataset, compared to 2.02 for DPM++2M, 1.97 for DPM2, and 11.90 for Heun for the same NFE. Meanwhile the Sampler Scheduler with the combined scheduling of SDE and ODE reaches 1.899, compared to 18.63 for Euler a, 3.14 for DPM2 a and 23.14 for DPM++ SDE.
\end{abstract}

\section{Introduction}
Diffusion models (DMs) \cite{ref1, ref2, ref3} have recently attracted great interest as powerful and expressive generative models. They have achieved unprecedented success in text-to-image synthesis and extended their impact to video generation\cite{ref4}, 3D modeling\cite{ref5}, speech synthesis\cite{ref6, ref7, ref8}, and lossless compression\cite{ref9} and other subject areas such as biology\cite{ref10}. However, the generation of DMs is slow due to the iterative noise removal process. While Generative Adversarial Networks (GANs) require only one network evaluation to generate a batch of images\cite{ref11}, DMs typically require hundreds or thousands of subsequent steps of Network Function Evaluation (NFE) to convert Gaussian noise into clean data\cite{ref1}. This inefficiency is becoming a critical obstacle in the adoption of DMs for downstream operations, showing an urgent need to accelerate the sampling process in DMs.\par  
One line of research aims to design generalized sampling techniques that can be easily applied to any pre-trained DM. Specifically, these techniques\cite{ref3, ref12, ref13, ref14, ref15, ref16, ref17, ref18} use an iterative backward process to gradually transform simple distributions into complex data distributions by solving differential equations. The associated vector fields (or drifts) that drive the evolution of the differential equations are predicted by the neural network. Prior samplers used to simulate these backward processes can be split into two categories: the ODE samplers\cite{ref12, ref16, ref17, ref18}, whose evolution beyond the initial randomization is deterministic, and the SDE samplers, whose generation trajectories are random. Research in this field has concentrated on building efficient numerical algorithms to speed up this process. Several publications illustrate that these samplers display their advantages in diverse scenarios.\par
In this study, we revisit and reevaluate the diffusion probabilistic flow ODE method, reveal its universal paradigm and extend it to the SDE method. We show that the entire sampling step is uncoupled, demonstrating the feasibility of using different sampling methods discretely. We emphasize the modular nature of the sampler to further optimize the design space for DM sampling. We introduce the Sampler Scheduler, an innovative approach that enables linear scheduling of different samplers. Extensive experiments are conducted to demonstrate its generality, superior efficiency and enhanced stability.\par

\section{Background}
\label{sec:Background}

\subsection{Diffusion: Noise Disturbance}

The diffusion model (DM)\cite{ref1, ref2, ref3, ref9} gradually adds Gaussian noise to the \(D\) dimensional random variables \(\boldsymbol{x}_0\in\mathbb{R}^D\) and perturbs the unknown data distribution \(q_0(\boldsymbol{x}_0)\) when \(t=0\) to a simple normal distribution \(q_T(\boldsymbol{x}_T)\approx\mathcal{N}(\boldsymbol{x}_T|\boldsymbol{O},\tilde{\sigma}^2\boldsymbol{I})\).\par

Transition distribution \(q_{t0}(\boldsymbol{x}_t|\boldsymbol{x}_0)\) at each moment in time \(t\in[0,T]\) satisfies

\begin{equation}
q_{t0}(\boldsymbol{x}_t|\boldsymbol{x}_0)=\mathcal{N}(\boldsymbol{x}_t|\alpha_t\boldsymbol{x}_0,\sigma_t^2\boldsymbol{I})
\label{eq:1}
\end{equation}

where \(\alpha_t,\sigma_t>0\) and the signal-to-noise ratio (SNR) \(\frac{a_{t^2}}{\sigma_{t^2}}\) is strictly decreasing with respect to \(t\).\par

Eq.\ref{eq:1} can be written as

\begin{equation}
\boldsymbol{x}_t=\alpha_t\boldsymbol{x}_0+\sigma_t\boldsymbol{\epsilon} 
\end{equation}

where \(\boldsymbol{\epsilon}{\sim}\mathcal{N}(\boldsymbol{O},\boldsymbol{I})\).

\subsection{Prediction Models: Noise Prediction and Data Prediction}

The diffusion model learns to recover data \(\boldsymbol{x}_0\) through a sequential denoising process, based on the noisy input \(\boldsymbol{x}_t\). . There are two alternative ways of defining the model.\par

Noise Prediction Model \(\epsilon_\theta(\boldsymbol{x}_t,t)\) attempts to predict noise based on data \(\boldsymbol{x}_t\). It optimizes the parameters \(\theta\) by the following objective\cite{ref2, ref3}

\begin{equation}
\min_\theta\mathbb{E}_{\boldsymbol{x_0}\sim q_0(\boldsymbol{x_0}),t\sim\mathcal{U}(0,1),\boldsymbol{\epsilon}\sim\mathcal{N}(\boldsymbol{O},\boldsymbol{I})}[\omega(t)\|\epsilon_\theta(\boldsymbol{x}_t,t)-\boldsymbol{\epsilon}\|_2^2]
\end{equation}

Data Predictive Model \(D_\theta(\boldsymbol{x}_t,t)\) tries to predict the denoised data \(\boldsymbol{x}_0\) based on data \(\boldsymbol{x}_t\). It optimizes the parameters \(\theta\) by the following objective\cite{ref19}

\begin{equation}
\min_\theta\mathbb{E}_{\boldsymbol{x_0}\sim q_0(\boldsymbol{x_0}),t\sim\mathcal{U}(0,1),\boldsymbol{\epsilon}\sim\mathcal{N}(\boldsymbol{O},\boldsymbol{I})}[\omega(t)\|D_\theta(\boldsymbol{x}_t,t)-\boldsymbol{x}_t\|_2^2]
\end{equation}

The relationship between the two is given by the following equation
\begin{equation}
\epsilon_\theta(\boldsymbol{x}_t,t)=\frac{\boldsymbol{x}_t-D_\theta(\boldsymbol{x}_t,t)}{\sigma_t}
\label{eq:5}
\end{equation}

For the sake of demonstrating convenience, we next illustrate using only the data prediction model \(D_\theta(\boldsymbol{x}_t,t)\). The noise prediction model  \(\epsilon_\theta(\boldsymbol{x}_t,t)\)  can be derived equivalently from the above transformations.

\subsection{Sampling: SDE and ODE}

In order to draw samples from the diffusion model, Song et al.\cite{ref3} proposed a backward synthesis process to solve the following stochastic differential equation (SDE) from \(\boldsymbol{x}_T{\sim}\mathcal{N}(\boldsymbol{O},{\sigma_T}^2\boldsymbol{I})\)

\begin{equation}
\mathrm{d}\boldsymbol{x}=-\underbrace{\dot{\sigma}(t)\sigma(t)\nabla_{\boldsymbol{x}}\log p(\boldsymbol{x};\sigma(t))\mathrm{d}t}_{\text{Probabilistic ODE}} - \underbrace { \beta _ t \sigma ( t ) ^ 2 \nabla _ { \boldsymbol { x }}\log p(\boldsymbol{x};\sigma(t))\mathrm{d}t+\sqrt{2\beta_t}\sigma(t)\mathrm{d}\omega_t}_{\text{Langevin process}}
\label{eq:6}
\end{equation}

where \(\nabla_{\boldsymbol{x}_t}\log p(\boldsymbol{x}_t;\sigma_t)\) is the score function (i.e., the gradient of the log probability), and \(\omega_{t}\) is the standard Wiener process, and \(\beta_{t}\) is a hyperparameter controlling the stochasticity of the process. The Langevin term can also be viewed as a combination of a deterministic score-based denoising term and a random noise injection term, whose net noise level contributions cancel each other out. Thus \(\beta_{t}\) effectively expresses the relative rate at which existing noise is replaced by new noise.\par

Thanks to the Tweedie formula , having the perfect noise reducer \(D_\theta\) is equivalent to having the score function:

\begin{equation}
D_\theta(\boldsymbol{x}_t,t)=\boldsymbol{x}_t+{\sigma_t}^2\nabla_{\boldsymbol{x}_t}\log p(\boldsymbol{x}_t;\sigma_t)
\end{equation}

Substituting into Eq.\ref{eq:6}, we get

\begin{equation}
d\boldsymbol{x}=\frac{\dot{\sigma}(t)}{\sigma(t)}\left(\boldsymbol{x}_t-D_\theta(\boldsymbol{x}_t,t)\right)dt+\beta_t(\boldsymbol{x}_t-D_\theta(\boldsymbol{x}_t,t))dt+\sqrt{2\beta_t}\sigma(t)d\omega_t
\end{equation}

When \(\beta=0\), the above equation degenerates to deterministic probabilistic flow ODE (deterministic probabilistic flow ODE)\cite{ref3}

\begin{equation}
d\boldsymbol{x}=\frac{\dot{\sigma}(t)}{\sigma(t)}(\boldsymbol{x}_t-D_\theta(\boldsymbol{x}_t,t))dt
\label{eq:9}
\end{equation}

\section{Sampler Scheduler}
\label{sec:Sampler Scheduler}

Even when using the same trained diffusion model, different solvers can lead to very different efficiencies and qualities of samplers. In this section, we present a unified formulation for the various probabilistic flow ODEs considered and show the feasibility of using different sampling methods discretely. We then extend the conclusions drawn to SDEs and establish a unified sampler scheduling framework that encompasses both ODE samplers and SDE samplers.

\subsection{Exogenous scaling and noise schedule}
Some approaches, such as variance-preserving diffusion models (VP)\cite{ref3} introduce additional scaling tables \(s(t)\) and consider \(x=s(t)x\) as a scaled version of the original variable \(\hat{x}\). This will change the time-dependent probability density of the backward process and hence the trajectory of the ODE solution. The resulting ODE is a generalization of Eq.\ref{eq:9}:

\begin{equation}
d\boldsymbol{x}=\frac{\dot{s}(t)}{s(t)}\frac{\dot{\sigma}(t)}{\sigma(t)}(\boldsymbol{x}_t-D_\theta(\boldsymbol{x}_t,t))dt
\end{equation}

We focus on non-scaling diffusion models in this paper. We explicitly undo the scaling of \(x\) in the model to keep the definition of \(p(\boldsymbol{x}_t;\sigma_t)\) independent of s(t) .\par

Karras et al.\cite{ref16} propose of the black-box nature of \(D_\theta\): the choices associated with the sampling process are largely independent of other components, such as network architecture and training details. In other words, the training process of \(D_\theta\) should not specify \(\sigma(t)\), \(s(t)\) and \(t_i\) and vice versa. Then the noise schedule \(\sigma_i=\sigma(t_i)\) as an exogenous component is low coupled to the sampler algorithm.

\subsection{Sampler Scheduler for ODE}
In the high-dimensional case, discretizing the SDE is often difficult and hard to converge in a few steps; in contrast, ODE sampling is easier to solve. We begin by checking ODE samplers, which provide a fruitful setting for analyzing sampling trajectories and their discretization. These insights carry over to the SDE, which we will reintroduce in the next section.

\begin{table}[htbp]
\centering
\begin{tabular}{l|l}
Sampler & Algorithm \\\hline
Euler & \(\boldsymbol{x}_{i+1}=\frac{\sigma_{i+1}}{\sigma_{i}}\boldsymbol{x}_{i}+\left(1-\frac{\sigma_{i+1}}{\sigma_{i}}\right)D_{\theta}(\boldsymbol{x}_{i})\) \\
Heun\cite{ref16} & \(\boldsymbol{x}_{i+1}=\frac{\sigma_{i+1}}{\sigma_i}\boldsymbol{x}_i+\left(1-\frac{\sigma_{i+1}}{\sigma_i}\right)D_\theta(\boldsymbol{x}_i)+\frac12\left(\frac{\sigma_i}{\sigma_{i+1}}-1\right)\left(D_\theta(\hat{\boldsymbol{x}}_{i+1})-D_\theta(\boldsymbol{x}_i)\right)\) \\
DPM++2M\cite{ref18} & \(\boldsymbol{x}_{i+1}=\frac{\sigma_{i+1}}{\sigma_i}\boldsymbol{x}_i+\left(1-\frac{\sigma_{i+1}}{\sigma_i}\right)D_\theta(\boldsymbol{x}_i)+\frac12\frac{\ln\frac{\sigma_{i-1}}{\sigma_i}}{\ln\frac{\sigma_i}{\sigma_{i+1}}}\left(D_\theta(\boldsymbol{x}_i)-D_\theta(\boldsymbol{x}_{i-1})\right)\) \\
DPM2\cite{ref17} & \(\boldsymbol{x}_{i+1}=\frac{\sigma_{i+1}}{\sigma_i}\boldsymbol{x}_i+\left(1-\frac{\sigma_{i+1}}{\sigma_i}\right)D_\theta(\boldsymbol{x}_i)+\frac{\sigma_i-\sigma_{i+1}}{\sqrt{\sigma_i\sigma_{i+1}}}\left(D_\theta\left(\hat{\boldsymbol{x}}_{i+\frac12}\right)-D_\theta(\boldsymbol{x}_i)\right)\)

\end{tabular}
\caption{\footnotesize
Iterative formulas for mainstream ODE samplers. This formula form is for formal briefness and ease of showing the commonality of different sampler algorithms, and the form is mathematically equivalent to the paradigm form given below. The Euler sampler uses only the prediction \(D_\theta(\boldsymbol{x}_i)\) of the current term \(\boldsymbol{x}_i\) as the model prediction, which is consistent with the basic ODE iterative formula. Heun is a second-order sampler, and its iterative formula actually adds a correction term to Euler's equation, using the prediction \(D_\theta(\hat{\boldsymbol{x}}_i)\) of the next step further back in the time series to correct the prediction \(D_\theta(\boldsymbol{x}_i)\) of the current step The correction term coefficient \(\frac{1}{2}\left(\frac{\sigma_i}{\sigma_{i+1}}-1\right)\) is an adjustment coefficient that varies with the noise table. The sum of the coefficients \(D_\theta(\boldsymbol{x}_i)\) and \(D_\theta(\hat{\boldsymbol{x}}_i)\) is zero, so the total amount of information and noise in the image does not change, but only the "direction" of denoising is corrected. DPM++ 2M, on the other hand, is a first-order sampler that uses the history term for correction, using the current step prediction \(D_\theta(\boldsymbol{x}_i)\) that is further back in the time series minus the previous step prediction \(D_\theta(\boldsymbol{x}_(i-1))\) for correction. DPM2 is also a second-order sampler that interpolates a geometric mean between the current step and the next step, correcting the current step prediction \(D_\theta(\boldsymbol{x}_i)\) by subtracting the prediction \(D_\theta\left(\hat{\boldsymbol{x}}_{i+\frac12}\right)\) of the next step further back in the time series.
}
\label{tabel:1}
\end{table}

For a given noise schedule \(\sigma_{i\in\{0,...,N\}}\), from Eq.\ref{eq:9} we can obtain the basic discretized ODE iteration formula

\begin{equation}
\boldsymbol{x}_{i+1}-\boldsymbol{x}_i=\frac{\sigma_{i+1}-\sigma_i}{\sigma_i}\big((\boldsymbol{x}_i-D_\theta(\boldsymbol{x}_i)\big)
\end{equation}

The formulas for reproducing the four ODE samplers in our framework are listed in Table \ref{tabel:1}. They were chosen because they are widely used and achieve state-of-the-art performance. Some of our formulations look quite different from the original paper, as indirection and recursion have been removed and default values in the code implementation are used for some optional hyperparameters. We take a "predictive-corrective" view of these mainstream methods, which provide a rich perspective on the use of historical terms or predicted future terms to correct \(D_\theta(\boldsymbol{x}_i)\). \par

We generalize the correction term to \(x_{j\in\{0,...,N\}}\) on each sampling step \(i\), and express the iterative formula as a linear combination of the current image \(\boldsymbol{x}_i\) with the prediction result \(\boldsymbol{D}_\theta(\boldsymbol{x})=\left(D_\theta(\boldsymbol{x}_0),D_\theta(\boldsymbol{x}_1),D_\theta(\boldsymbol{x}_2),\cdots,D_\theta(\boldsymbol{x}_N)\right)\) of all images \(\boldsymbol{x}\) on the time series in the time series.

\begin{equation}
\boldsymbol{x}_{i+1}=A\boldsymbol{x}_i+\boldsymbol{B}\boldsymbol{D}_{\theta}(x)
\label{eq:12}
\end{equation}

We decompose \(\boldsymbol{B}\) into \(\boldsymbol{B}=B\boldsymbol{c}\) where \(\sum c_j=1\). Then Eq.\ref{eq:12} becomes

\begin{equation}
\boldsymbol{x}_{i+1}=A\boldsymbol{x}_i+B\boldsymbol{c}\boldsymbol{D}_{\theta}(x)
\end{equation}

For general image data, on each channel, the value of each pixel point is distributed in [0,255]. To avoid overflow of this value during the update process at each sampling step, we define the total information function \(S(x)\) that satisfies \(S(\boldsymbol{x})=S(D_{\theta}(\boldsymbol{x}))\), regardless of the degree of noise within the image. Also, to make the iteration process conform to the noise schedule, we define the noise amount function \(N(x)\) that satisfies \(N(\boldsymbol{x}_i)=\sigma_i,N(D_\theta(\boldsymbol{x}_i))=0\).\par

Then we can get the set of iterative equations for discrete sampling steps defined in terms of total information and noise:

\begin{equation}
\left.\left\{\begin{aligned}S(\boldsymbol{x}_{i+1})&=A\cdot S(\boldsymbol{x}_i)+B\cdot \boldsymbol{c}\boldsymbol{S}(D_{\theta}(\boldsymbol{x}))\\N(\boldsymbol{x}_{i+1})&=A\cdot N(\boldsymbol{x}_i)+B\cdot \boldsymbol{c}\boldsymbol{N}(D_{\theta}(\boldsymbol{x}))\end{aligned}\right.\right.
\end{equation}

Substitute the corresponding values and solve the equation

\begin{equation}
\begin{bmatrix}
1&1\\\sigma_i&0\end{bmatrix}\begin{bmatrix}A\\B\end{bmatrix}=\begin{bmatrix}1\\\sigma_{i+1}
\end{bmatrix}
\end{equation}

The coefficients A, B have unique solution

\begin{equation}
\begin{cases}\quad A=\dfrac{\sigma_{i+1}}{\sigma_i}\\B=1-\dfrac{\sigma_{i+1}}{\sigma_i}
\end{cases}
\end{equation}

We obtain the general noise reduction formula for discrete ODEs

\begin{equation}
\begin{gathered}
\boldsymbol{x}_{i+1}=\frac{\sigma_{i+1}}{\sigma_{i}}\boldsymbol{x}_{i}+\left(1-\frac{\sigma_{i+1}}{\sigma_{i}}\right)\boldsymbol{c}\boldsymbol{D}_{\boldsymbol{\theta}}(\boldsymbol{x}) \\
=\frac{\sigma_{i+1}}{\sigma_i}\boldsymbol{x}_i+\left(1-\frac{\sigma_{i+1}}{\sigma_i}\right)\sum c_jD_\theta(\boldsymbol{x}_j) 
\end{gathered}
\end{equation}

where \(\boldsymbol{x}_j=\begin{cases}\boldsymbol{x}_j,~j\leq i\\\hat{\boldsymbol{x}}_j,~j>i&\end{cases}\)

For the history step, \(\boldsymbol{x}_{j\leq i}\) is a definite obtained value. For the future step, \(\boldsymbol{x}_{j>i}\) is the baseline estimate obtained by the Euler sampling method

\begin{equation}
\hat{\boldsymbol{x}}_{i+1}=-\frac{\sigma_{i+1}}{\sigma_i}\boldsymbol{x}_i+\left(1-\frac{\sigma_{i+1}}{\sigma_i}\right)D_\theta(\boldsymbol{x}_i)
\end{equation}

Consider the case of interpolation. There is an interpolation \(\boldsymbol{x}_{i+k}=f(\boldsymbol{x}_i,\boldsymbol{x}_{i+1})\) between \(\boldsymbol{x}_i\) and \(\boldsymbol{x}_{i+1}\), whose corresponding noise \(\sigma_{i+k}=f(\sigma_i,\sigma_{i+1})\). We obtain a more general equation

\begin{equation}
\hat{\boldsymbol{x}}_{i+k}=\frac{\sigma_{i+k}}{\sigma_i}\boldsymbol{x}_i+\left(1-\frac{\sigma_{i+k}}{\sigma_i}\right)D_\theta(\boldsymbol{x}_i),k\in(0,1]
\end{equation}

The only coefficient that affects the noise reduction effect is \(\boldsymbol{c}\boldsymbol{D}_\theta(x)\), and the only variable \(\boldsymbol{c}(i,\sigma)\) determines the performance of the ODE sampler.\par
We note that previous work has attempted to find an optimal global sampler to use throughout the sampling process. For sampling processes that use a single global sampler throughout, \(\boldsymbol{c}(i,\sigma)\) is only relevant with the noise schedule \(\sigma\)

\begin{equation}
\boldsymbol{c}(i,\sigma)=\boldsymbol{c}(\sigma)
\end{equation}

This parameter is used consistently i.e. \(\boldsymbol{c}\) is uncorrelated with \(i\) for any given moment \(i\). We list the values of \(\boldsymbol{c}\) for common ODE samplers in Appendix \ref{app:c}.\par

We believe that using static \(\boldsymbol{c}(\sigma)\) on all sampling steps is unnecessary. In our framework, there are no implicit dependencies between each sampling step - in principle any choice of \(\boldsymbol{c}(i,\sigma)\) (within reason) produces a functional model, and discretely using different samplers for different sampling steps will provide greater flexibility for the sampling process to change or even optimize the trajectory of the solution. Compared to \(\sigma\), our work is more concerned with the relation \(\boldsymbol{c}(i)\) between \(\boldsymbol{c}\) and \(i\) after the introduction of \(i\).\par 

Of course, in practice, certain choices and combinations will work better than others. Each sampler has been theoretically proven to have unique advantages, we compose \(\boldsymbol{c}(i)\) based on the better solutions \(\boldsymbol{c}_{Euler}(\sigma),\boldsymbol{c}_{DPM++}(\sigma),...\) obtained from previous work.\par

Accordingly, we propose a sampling scheduler to adjust the ODE sampling method according to the i dynamically.

\begin{equation}
\boldsymbol{x}_{i+1}=\frac{\sigma_{i+1}}{\sigma_i}\boldsymbol{x}_i+\left(1-\frac{\sigma_{i+1}}{\sigma_i}\right)\boldsymbol{c}(i)\boldsymbol{D}_\theta(\boldsymbol{x})
\end{equation}
Simply, we can use multiple ODE samplers for linear step-by-step scheduling based on existing ODE samplers. Such a \(\boldsymbol{c}(i)\) theoretically achieves better performance than a single global \(\boldsymbol{c}\).

\begin{algorithm} [htbp]
    \caption{ODE Samplers Scheduler} 
    \label{alg:1} 
    \begin{algorithmic}
    \REQUIRE \(D_\theta(x),~\sigma_{i\in\{0,...,N\}},~c_{k\in\{1,...,M\},i}\mathrm{and~corresponding~ }s_k,\Sigma_{k=1}^Ms_k=N)\)
    \STATE \(\mathrm{sample~}\boldsymbol{x}_0{\sim}\mathcal{N}(0,\sigma^2)\)
    \FOR{\(k\in\{1,\ldots,M\}\)}
        \FOR{\(j\in\{1,\ldots,s_k\}\)}
        \STATE \(\boldsymbol{x}_{i+1}\leftarrow\frac{\sigma_{i+1}}{\sigma_i}\boldsymbol{x}_i+\left(1-\frac{\sigma_{i+1}}{\sigma_i}\right)\sum_{i=0}^Nc_{k,i}D_\theta(x_i)\)
        \ENDFOR
    \ENDFOR
    \RETURN \(\boldsymbol{x}_N\)
    \end{algorithmic} 
\end{algorithm}

\subsection{Sampler Scheduler for SDE}
ODE-based deterministic sampling has many benefits, such as the ability to transform a real image into its corresponding latent representation by inverting the ODE. However, it tends to result in poorer output quality than random sampling that injects new noise into the image at each step. Song Yang et al.\cite{ref3} explain this score-matching trap from the perspective of high-dimensional data distributions-estimating the score function is inaccurate in low-density regions, where there are few data points available to compute score matching targets are few data points. Otherwise, empirically, adding new noise at each step is a more favorable choice for stabilizing the neural network. Otherwise the results will be closer to the average of the original training dataset than to the outcome.\par
The stochastic differential equation in Eq.\ref{eq:6} i.e. can be viewed as a probabilistic flow ODE plus a corresponding a Langevin process. Implicit Langevin diffusion pushes the samples towards the desired marginal distribution at a given time, thus actively correcting errors that occur in earlier sampling steps. Thus SDE samplers are often considered to produce higher quality images.

\begin{figure}[htbp]
  \centering
  \includegraphics[width=0.6\textwidth]{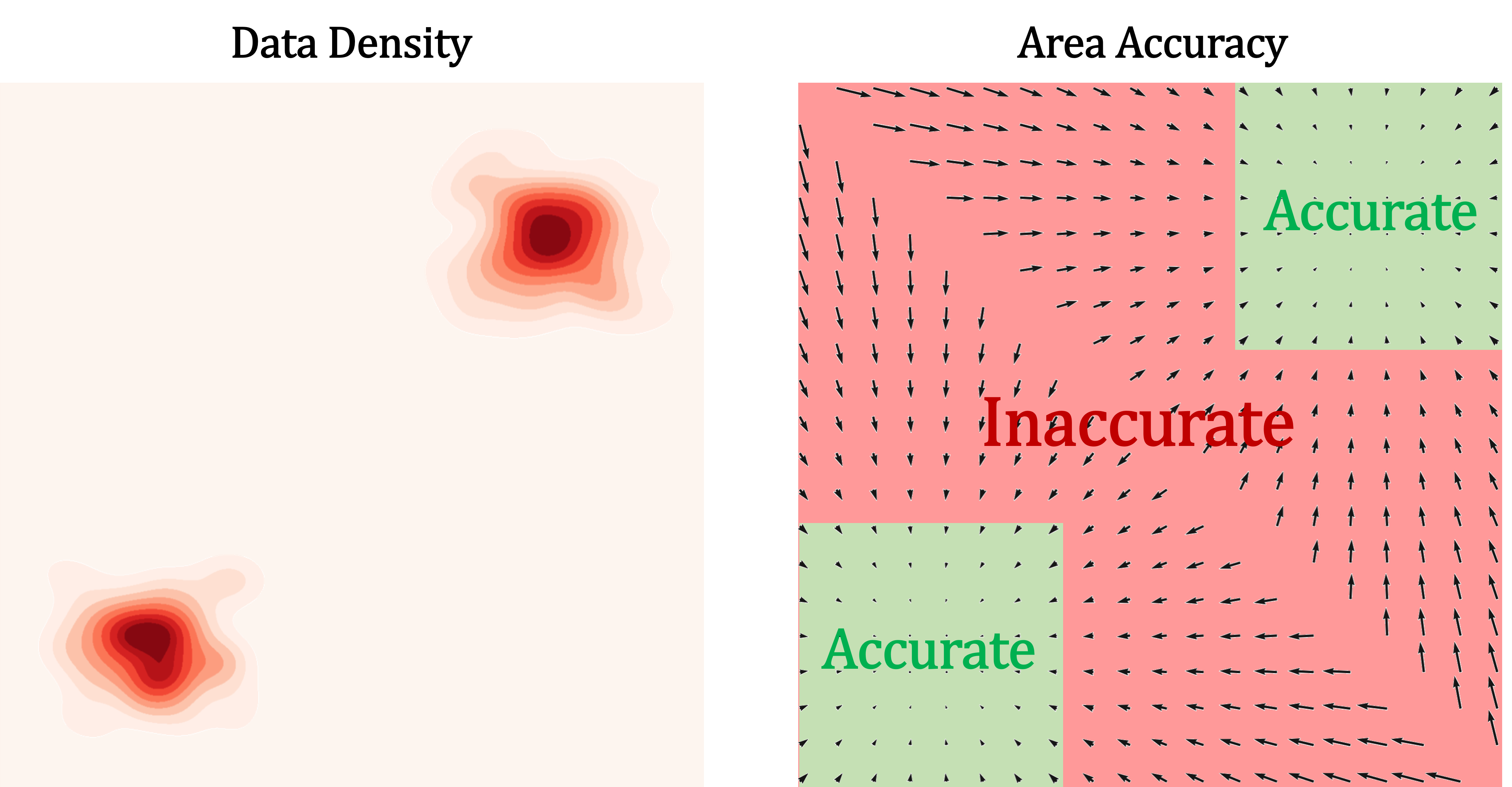}
  \caption{Data Density and Area Accuracy.}
  \label{fig:1}
\end{figure}

\begin{table}[htbp]
\footnotesize
\centering
\begin{tabular}{l|l}
Sampler & Algorithm \\\hline
Euler a\cite{ref16} & \(\boldsymbol{x}_{i+1}=\frac{\sigma_{i+1}^2}{\sigma_i^2}\boldsymbol{x}_i+\left(1-\frac{\sigma_{i+1}^2}{\sigma_i^2}\right)D_\theta(\boldsymbol{x}_i)+\frac{\sigma_{i+1}}{\sigma_i}\sqrt{\sigma_i^2-\sigma_{i+1}^2}\boldsymbol{\epsilon}_i\) \\
DPM++ SDE\cite{ref18} & \(\boldsymbol{x}_{i+1}=\frac{\sigma_{i+1}^2}{{\sigma_i}^2}\boldsymbol{x}_i+\left(1-\frac{\sigma_{i+1}^2}{{\sigma_i}^2}\right)D_\theta(\boldsymbol{x}_i)+\frac12\left(1-\frac{\sigma_{i+1}^2}{{\sigma_i}^2}\right)\left(D_\theta(\hat{\boldsymbol{x}}_{i+1})-D_\theta(\pmb{x}_i)\right)+\frac{\sigma_{i+1}}{\sigma_i}\sqrt{\sigma_i^2-{\sigma_{i+1}}^2}\boldsymbol{\epsilon_i}\) \\
DPM++ 2S a\cite{ref18} & \(\boldsymbol{x}_{i+1}=\frac{\sigma_{i+1}^2}{{\sigma_i}^2}\boldsymbol{x}_i+\left(1-\frac{{\sigma_{i+1}}^2}{{\sigma_i}^2}\right)D_\theta(\hat{\boldsymbol{x}}_{i+1})+\frac{\sigma_{i+1}}{\sigma_i}\sqrt{{\sigma_i}^2-{\sigma_{i+1}}^2}\boldsymbol{\epsilon}_i\) \\
DPM++ 2M SDE\cite{ref18} & \(\boldsymbol{x}_{i+1}= \scriptsize{\frac{{\sigma_{i+1}}^2}{{\sigma_i}^2}\boldsymbol{x}_i+\left(1-\frac{{\sigma_{i+1}}^2}{{\sigma_i}^2}\right)D_\theta({\boldsymbol{x}}_i)+\frac12\left(1-\frac{{\sigma_{i+1}}^2}{{\sigma_i}^2}\right)\frac{\ln\frac{\sigma_i}{\sigma_{i+1}}}{\ln\frac{\sigma_{i-1}}{\sigma_i}}\left(D_\theta({\boldsymbol{x}}_i)-D_\theta({\boldsymbol{x}}_{i-1})\right)+\frac{\sigma_{i+1}}{\sigma_i}\sqrt{\sigma_i^2-{\sigma_{i+1}}^2}\boldsymbol{\epsilon_i}}\)
 \\
DPM2 a\cite{ref17} & \(\boldsymbol{x}_{i+1}=\frac{\sigma_{i+1}{}^2}{\sigma_{i}{}^2}\boldsymbol{x}_i+\left(1-\frac{\sigma_{i+1}{}^2}{\sigma_{i}{}^2}\right)D_\theta(\boldsymbol{x}_i)+\frac{\sigma_{i}{}^2-\sigma_{i+1}{}^2}{\sigma_i\sigma_{i+1}{}^2}\left(D_\theta\left(\hat{\boldsymbol{x}}_{i+\frac12}\right)-D_\theta(\boldsymbol{x}_i)\right)+\frac{\sigma_{i+1}}{\sigma_i}\sqrt{\sigma_i^2-{\sigma_{i+1}}^2}\boldsymbol{\epsilon_i}\)

\end{tabular}
\caption{\footnotesize
Iterative formulas for mainstream SDE samplers. This form of the formulas is intended to be formally brief and to facilitate the demonstration of the commonalities of the different sampler algorithms, and the form is mathematically equivalent to the paradigm form given below. Euler a sampler adds ancestor sampling to the Euler sampler. Similarly to Euler, Euler a takes only the current term prediction \(D_\theta(\boldsymbol{x}_i)\). DPM++SDE is a second-order sampler, similar to Heun and Euler, and its update formula actually adds a correction term to the Euler a equation, using the prediction \(D_\theta(\hat{\boldsymbol{x}}_{i+1})\) of the next step further back in the time series to correct the prediction \(D_\theta(\boldsymbol{x}_i)\) of the current step. Note that in crowsonkb's kdiffusion\cite{ref25} implementation, which is widely used by the community, the default correction term coefficients are twice as high as those listed here, namely \(\begin{pmatrix}\frac{\sigma_i}{\sigma_{i+1}}-1\end{pmatrix}\), which degrades the update formula for DPM++ SDE to DPM++ 2S a. For DPM++ 2S a, the actual computation is \(D_\theta(\boldsymbol{x}_i)\) is eliminated and only the next Euler-based prediction \(D_\theta(\hat{\boldsymbol{x}}_{i+1})\) is used for each prediction. Similarly to DPM++ 2M, DPM++ 2M SDE uses the current step prediction \(D_\theta(\boldsymbol{x}_i)\) that is further back in the time series minus the prediction \(D_\theta(\boldsymbol{x}_{i-1})\) of the previous step to perform the correction. Similarly to DPM2, DPM2 a uses the prediction \(D_\theta\left(\hat{\boldsymbol{x}}_{i+\frac12}\right)\) of the next step further back in the time series to correct the prediction \(D_\theta(\boldsymbol{x}_i)\) of the current step . We consider the algorithm with ancestor sampling (a) as a tail and the algorithm with stochastic differential equations (SDE) as a tail to be of the same type. It is not difficult to find a close connection between the ODE and SDE versions of the same algorithm as well as consistent variations in parameter differences.
}
\label{tabel:2}
\end{table}

 The formulas of the five SDE samplers in our framework are listed in Table \ref{tabel:2}. They are chosen because they are widely used and achieve state-of-the-art performance. Some of our formulas look quite different from the original paper, as indirection and recursion have been removed and default values in the code implementation are used for some optional hyperparameters. We list \(\boldsymbol{c}\) of common SDE sampler's in Appendix \ref{app:c}\par
Similar to ODE, we give a general iterative formulation of the SDE sampling algorithm. Detailed derivation is in Appendix \ref{app:b}.

\begin{equation}
\boldsymbol{x}_{i+1}=\frac{\sigma_{i+1}^2}{\sigma_i^2}\boldsymbol{x}_i+\left(1-\frac{\sigma_{i+1}^2}{\sigma_i^2}\right)\boldsymbol{c}(i)\boldsymbol{D}_\theta(\boldsymbol{x}_i)+\frac{\sigma_{i+1}}{\sigma_i}\sqrt{\sigma_i^2-\sigma_{i+1}^2}\boldsymbol{\epsilon}_i
\end{equation}

where \(\left.\boldsymbol{x}_j=\left\{\begin{matrix}\boldsymbol{x}_j,&j\leq i\\\hat{\boldsymbol{x}}_j,&j>i\end{matrix}\right.\right.,\quad\hat{\boldsymbol{x}}_{i+k}=\frac{\sigma_{i+k}}{\sigma_i}\boldsymbol{x}_i+\left(1-\frac{\sigma_{i+k}}{\sigma_i}\right)D_\theta(\boldsymbol{x}_i),k\in(0,1]\mathrm{~and~}\forall i,\sum c_j(i)=1\)

\subsection{Put ODE and SDE together}

Adding randomness can be effective in correcting errors caused by earlier sampling steps, but it has its own drawbacks. Karras et al.\cite{ref16} argued that excessive Langevin-like addition and removal of noise leads to gradual loss of detail in the generated images with all datasets and denoiser networks. And there is a tendency for colors to become oversaturated at very low noise levels. This led Karras et al.\cite{ref16} to propose the strategy of limiting the total amount of randomness and increasing the standard deviation of the newly added noise, but it is not a complete solution to this inherent problem related to the SDE algorithm itself.\par
We believe that discretely employing the SDE and ODE sampler algorithms separately at different sampling steps can further alleviate this problem. Specifically, for the early sampling step, we prefer to use the SDE sampler to push the samples away from the low-density region, while for the late sampling step, we prefer to use the ODE sampler to quickly converge to the sampling results while preserving the details with reduced drift.\par

\begin{figure}[!h]
  \centering
  \includegraphics[width=0.6\textwidth]{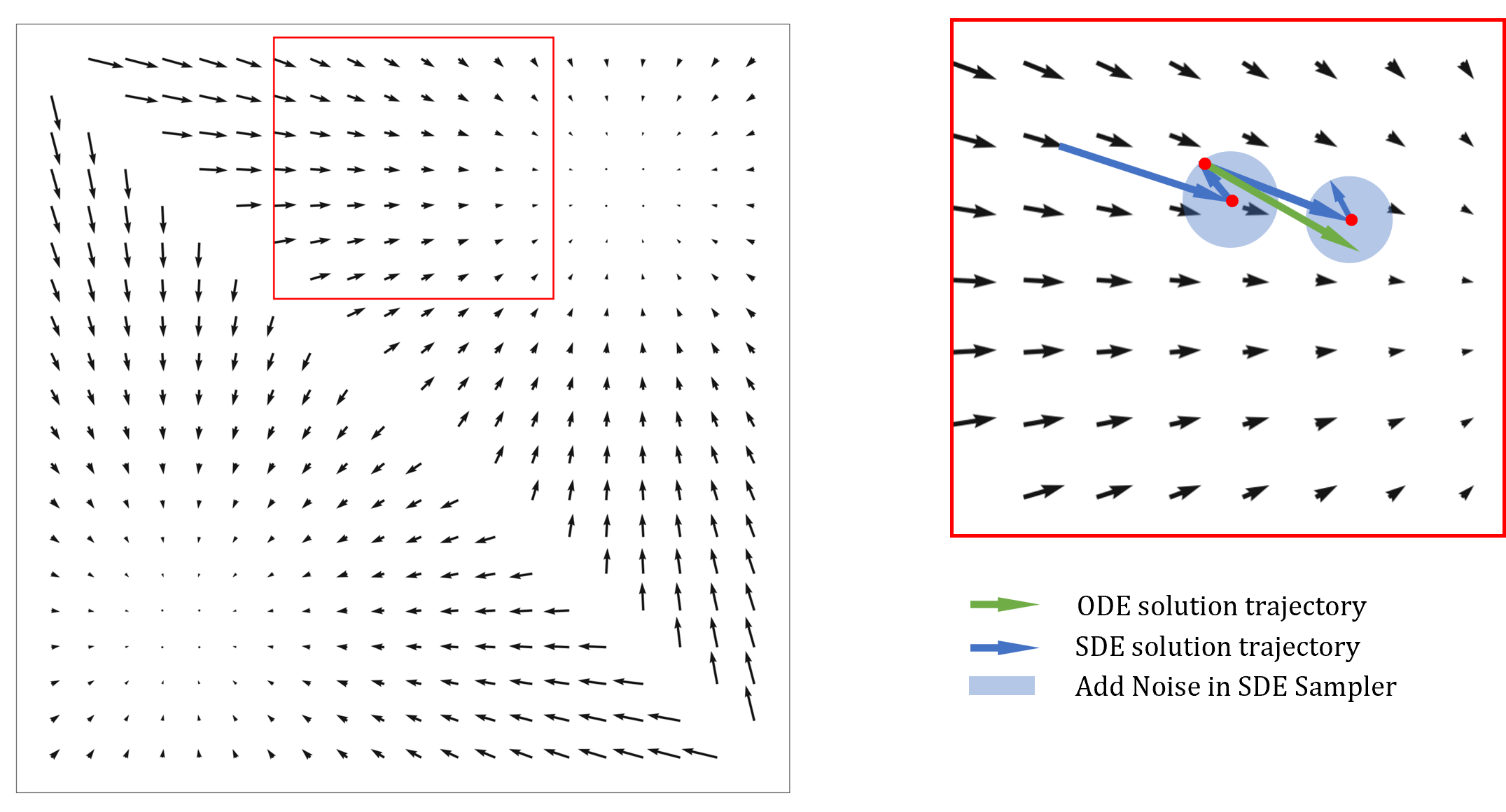}
  \caption{In the later sampling step, using the ODE sampler can quickly converge to the sampling result while retaining details and reducing drift.}
  \label{fig:2}
\end{figure}

Further, we integrated both ODE sampler and SDE sampler into our Samplers Scheduler. At the same time, we follow the modular design of sampler independence. Except for necessary additional parameters, such as the settings of different sampling algorithms to be used on different sampling steps, our Samplers Scheduler can be encapsulated as a single sampler function, just like any global sampler, without destroying or rewriting any of the existing design space. We demonstrate this by implementing the design as a plug-in extension to Stable Diffusion WebUI\cite{ref21}, with code open sourced from \href{https://github.com/Carzit/sd-webui-samplers-scheduler}{https://github.com/Carzit/sd-webui-samplers-scheduler}.

\begin{algorithm} [!h]
    \caption{Samplers Scheduler} 
    \label{alg:2} 
    \begin{algorithmic}
    \REQUIRE \(D_\theta(\boldsymbol{x}),~\sigma_i,~Sampler_k,~steps_k,~i\in\{0,...N\},k\in\{0,...,M\}\)
    \STATE \(\mathrm{sample~}\boldsymbol{x}_0{\sim}\mathcal{N}(0,\sigma^2)\)
    \FOR{\(k\in\{1,\ldots,M\}\)}
        \STATE \(\sigma_{max}=\sigma_{\Sigma_{j<k}steps_j},\sigma_{min}=\sigma_{\Sigma_{j\leq k}steps_j-1}\)
        \STATE \(\{\sigma_{k,0},\sigma_{k,2},...,\sigma_{k,\textit{steps}_k}\}=NoiseSchedule(steps_k,\sigma_{max},\sigma_{min})\)
        \STATE \(\boldsymbol{x}_{k,0}=\boldsymbol{x}_{k-1,steps_{k-1}}\)
    
        \FOR{\(j\in\{1,\ldots,steps_k\}\)}
        \STATE \(\boldsymbol{x}_{k,j}=Sampler_k(D_\theta(\boldsymbol{x}),\boldsymbol{x}_{k,j-1},\sigma_{k,j-1})\)
        \ENDFOR
    \ENDFOR
    \STATE \(\boldsymbol{x}_N=\boldsymbol{x}_{M,steps_M}\)
    \RETURN \(\boldsymbol{x}_N\)
    \end{algorithmic} 
\end{algorithm}

\section{Experiments}
\subsection{Experiments on Standard Benchmarks}

We use the Fréchet Inception Distance (FID)\cite{ref22} to further investigate image sampling quality. To evaluate the balance between sample quality and inference speed, we report the FID score and the number of function evaluations (NFE) for the sampler. We borrow the VP/EDM model pre-trained in EDM\cite{ref23} on the CIFAR-10\cite{ref24} dataset and use the code implementation in k-diffusion\cite{ref25} for each sampler.

\begin{figure}[htbp]
	\centering
	\subfloat[ODE]{\includegraphics[width=.48\columnwidth]{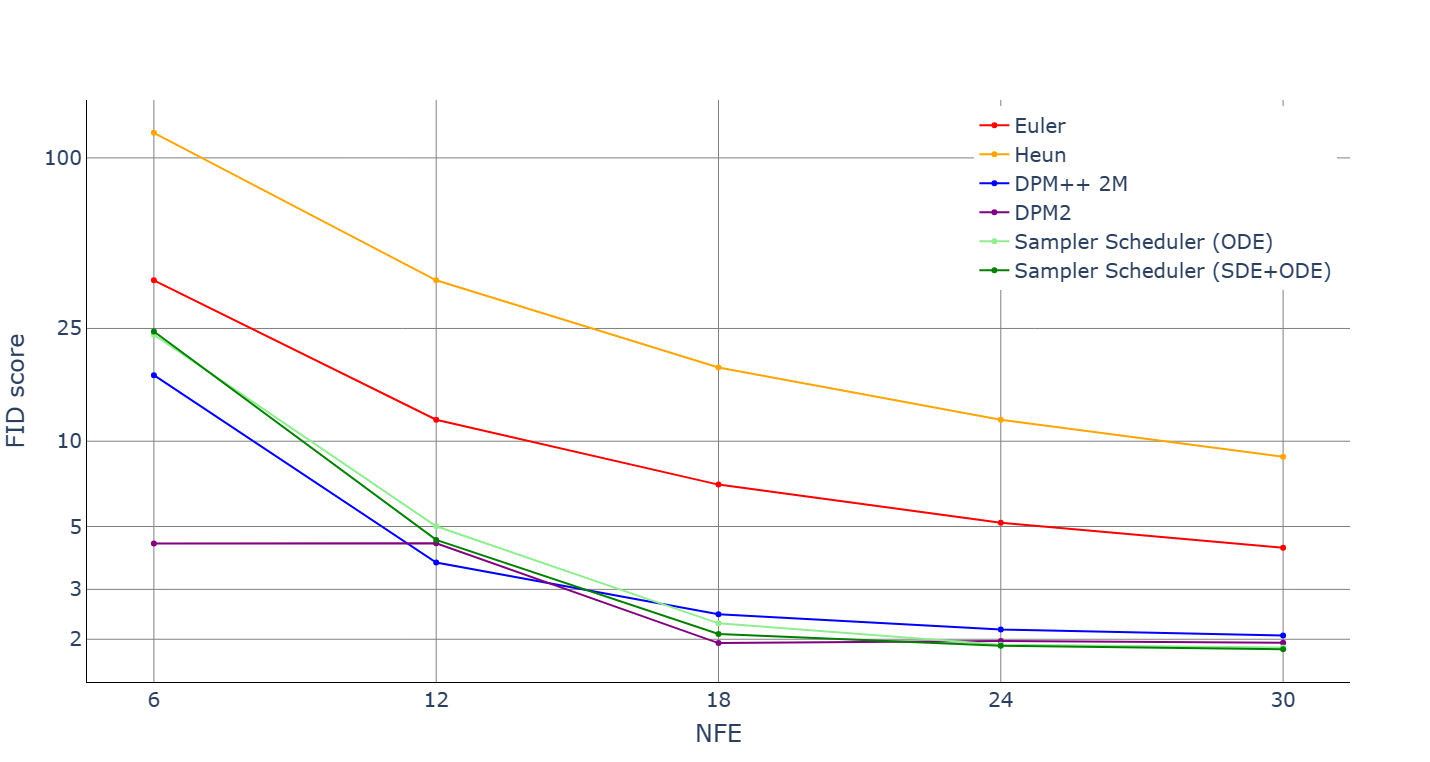}}
	\subfloat[SDE]{\includegraphics[width=.48\columnwidth]{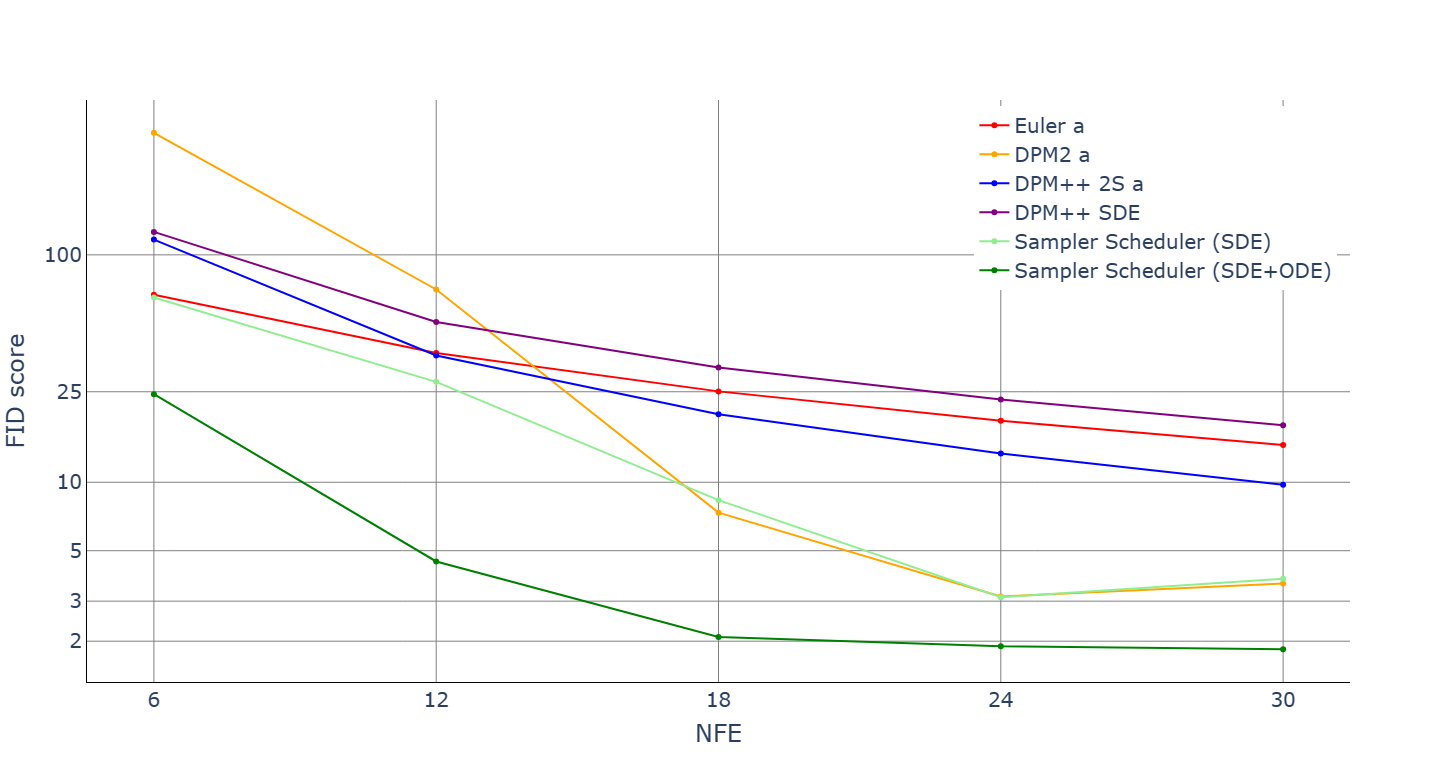}}
	\caption{Comparison among various samplers on pretrained model.}
\label{fig:3}
\end{figure}

We compare the first-order approach proposed in our Sampler Scheduler and EDM with the second-order approach proposed in DPM(++) while controlling the same NFE. We report in Fig. 3 how the FID scores on CIFAR-10 compare to the NFE.\par
For ODE samplers, the Sampler Scheduler that dynamically schedules these ODE samplers achieves better FID scores at NFE > 24 compared to other single global ODE samplers. Meanwhile, the Sampler Scheduler that uses the combined scheduling approach of applying SDE samplers on early sampling steps and applying ODEs on late sampling steps that we suggested in Section 3.4 also obtains similarly better scores at NFE of 24 (1.899 vs. 1.914).\par
For SDE, our scheduling configurations are categorized into two types: a Sampler Scheduler consisting of different SDE samplers and a Comprehensive Sampler Scheduler as proposed in Section 3.4.Our Comprehensive Sampler Scheduler achieves better FID scores across all NFEs when compared to the respective SDE samplers. We note that at NFE = 12, the Integrated Sampler Scheduler achieves an 87.58\% improvement in FID compared to the single-step DPM-Solver++ (36.100 vs. 4.484).

\subsection{Experiments on Large-scale Text-to-image Model}

We note that Stable Diffusion\cite{ref26} has been widely noticed and applied in the field of image generation from diffusion models. Therefore, we further apply the Sampler Scheduler to the text-to-image Stable Diffusion v1.5\cite{ref28} pre-trained on LAION-5B\cite{ref27} at 512×512 resolution. We employ the commonly used classifier-guidance\cite{ref29, ref30} for sampling and use the COCO\cite{ref31} validation set for evaluation, where we use the open-source ViT-g/14\cite{ref32} to evaluate CLIP\cite{ref33} scores for text-image alignment and the open-source LAION-Aesthetic Predictor\cite{ref34} to evaluate the Aesthetic Score for visual quality.\par
We use 5K randomly sampled words from the validation set for prompt generation, compute all evaluation metrics, and plot CLIP versus aesthetic score trade-off curves, where the unclassifier-guided weights w are \{2, 3, 5, 8\}.

\begin{figure}[htbp]
\centering
{\includegraphics[width=.8\columnwidth]{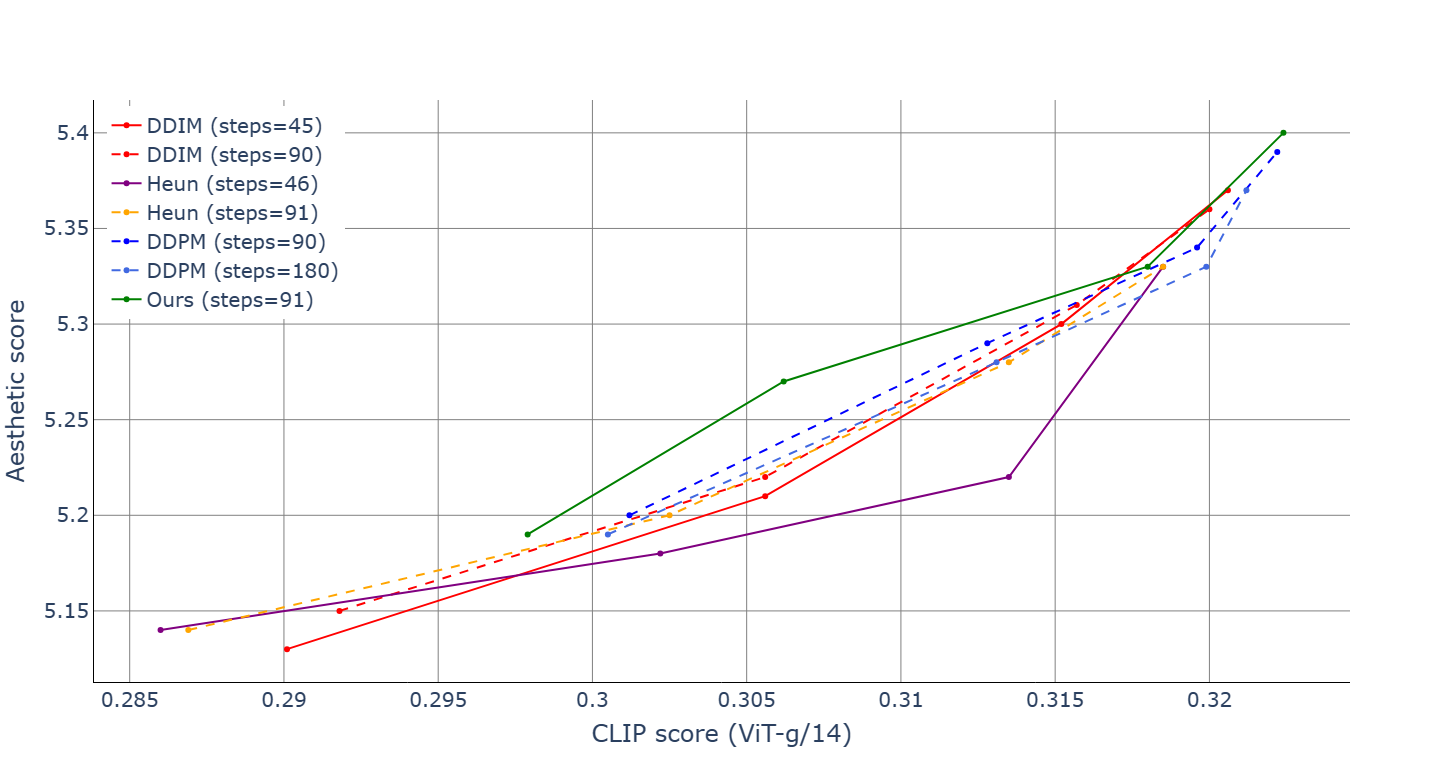}}
\caption{CLIP ViT-g/14 score versus Aesthetic score for text-to-image
generation at 512×512 resolution, using Stable Diffusion v1.5 with a varying classifier-free guidance
weight w = \{2, 3, 5, 8\}.}
\label{fig:4}
\end{figure}

We compare with the commonly used ODE sampler DDIM\cite{ref12} and the SDE sampler DDPM\cite{ref1}. For the Sampler Scheduler, we use a 31-step Heun and 60-step DPM++2M stepwise scheduling using 121 NFE(i.e., 91 sampling steps). As shown in Fig.\ref{fig:4}, Sampler Scheduler achieves better Clip scores in most cases, given the same unclassified classifier guidance weights. And with similar Clip scores, Sampler Scheduler usually achieves better visual quality (aesthetic score). Notably, when the CLIP scores are high, DDPM and DDIM usually obtain similar performance to Sampler Scheduler in terms of visual quality, while Sampler Scheduler is more time-efficient. These findings suggest that the Sampler Scheduler balances text-image alignment (CLIP score) or visual quality (aesthetic score) more efficiently than previous samplers.

\section{Conclusion and Limitations}
In this paper, we introduce the diffusion model and Sampler Scheduler for generative processes involving differential equations. By scheduling different samplers on different sampling steps, Sampler Scheduler exploits and even enhances the respective advantages of the sampling algorithms of ODE and SDE. Theoretically, Sampler Scheduler can achieve better sample quality. Empirically, Sampler Scheduler achieves an excellent balance between quality and time and improves text-image alignment/visual quality in text-to-image Stable Diffusion models.\par
We currently adopt the better global sampler algorithm c obtained in previous work to constitute \(\boldsymbol{c}(i)\) of the selection. In the future, better algorithms can also be included, so our framework is growable. In the meantime we anticipate developing a broader approach to scheduling than the current single linear scheduling, such as combining multiple samplers in parallel to try more \(\boldsymbol{c}(i)\) to realize the full potential of the Sampler Scheduler framework.\par
The current limitation of the Sampler Scheduler is the lack of a general approach for hyperparameter selection. To some extent, the DPM network parameters \(\theta\) are correlated \(\boldsymbol{c}(i)\). Our approach yields better solutions than the existing single global sampler, but it is difficult to obtain a generalized hyperparameter configuration that is theoretically optimal for a wide range of models. As a future direction, we will further develop a principled method to automatically select the optimal hyperparameters of the Sampler Scheduler based on the error analysis of the model.

\clearpage
\bibliographystyle{plain}  
\bibliography{references}  

\clearpage

\appendix
{\LARGE Appendix}
\section{Related Works and Discussion}
In order to unleash the extraordinary generative potential of diffusion models and make them widely available, a large number of research efforts have been devoted to improving the quality of the images they generate and achieving accelerated sampling. In addition to the work of Song eta al.\cite{ref3}, Karras et al.\cite{ref16}, and Lu et al.\cite{ref17, ref18}, which we have mentioned or cited above, we are also concerned with some other recent related work.\par
We note that RES\cite{ref35} is a concurrent work (concurrent work) that uses an improved exponential integrator to further reduce the sampler error range. Its basically available as a DPM++ 2M SDE implementation using Heun Solver and added to our Sampler Scheduler framework.\par
We note that UniPC\cite{ref36} is a concurrent work that implements a unified predictive corrector framework called UniPC by developing the Unified Corrector (UniC), which demonstrates excellent performance in very few steps of fast sampling, but has limited effect in improving the quality of the generated image at generally more than 30 steps. Empirically, its use on early sampling steps in the Sampler Scheduler framework accelerates the convergence rate, but the improvement in image quality is subject to further experiments with different scheduling configurations.\par
We note that Restart\cite{ref37} is a concurrent effort that exploits and even enhances the respective advantages of ODE and SDE by combining an additional forward step that adds a significant amount of noise with the corresponding backward ODE. This is in line with the idea that Sampler Scheduler exploits the respective advantages of ODE and SDE by scheduling the corresponding ODE and SDE samplers on different off-walks. Unlike our work, Restart adds a large amount of noise at once in the middle of a sample to achieve fallback across sampling steps.

\section{Discretized Iterative Formula for SDE}
\label{app:b}

Consider a range of noise scales \(\sigma_0<\sigma_1<\sigma_2<\cdotp\cdotp<\sigma_N\). By sequentially perturbing the data points \(\boldsymbol{x}_0\) with these noise scales, we obtain a Markov chain \(\boldsymbol{x}_0\to\boldsymbol{x}_1\to \boldsymbol{x}_2\to\cdots\to \boldsymbol{x}_N\) where

\begin{equation}
p(\boldsymbol{x}_i|\boldsymbol{x}_{i-1})=\mathcal{N}(\boldsymbol{x}_i;\boldsymbol{x}_{i-1},({\sigma_i}^2-{\sigma_{i-1}}^2)\boldsymbol{I}),i=1,2,\cdots,N
\end{equation}

If we parameterize the reverse transfer kernel as \(p_\theta(\boldsymbol{x}_{i-1}|\boldsymbol{x}_i)=\mathcal{N}(\boldsymbol{x}_{i-1};\mu_\theta(\boldsymbol{x}_i,i),\tau_i^2\boldsymbol{I})\), then

\begin{equation}
\begin{gathered}
L_{t-1}=\mathbb{E}_q\big[D_{KL}\big(q(\boldsymbol{x}_{i-1}|\boldsymbol{x}_i,\boldsymbol{x}_0)\big)\|p_\theta(\boldsymbol{x}_{i-1}|\boldsymbol{x}_i)\big]\\
=\mathbb{E}_q\left[\frac{1}{2{\tau_i}^2}\left\|\frac{\sigma_{i-1}^2}{{\sigma_i}^2}\boldsymbol{x}_i+\left(1-\frac{\sigma_{i-1}^2}{{\sigma_i}^2}\right)\boldsymbol{x}_0-\mu_\theta(\boldsymbol{x}_i,i)\right\|_2^2\right]+C\\
=\mathbb{E}_{\boldsymbol{x}_{0},\boldsymbol{\epsilon}}\left[\frac{1}{2{\tau_{i}}^{2}}\left\|\boldsymbol{x}_{i}(\boldsymbol{x}_{0},\boldsymbol{\epsilon})+\frac{{\sigma_{i}}^{2}-{\sigma_{i-1}}^{2}}{\sigma_{i}}\boldsymbol{\epsilon}-\mu_{\theta}(\boldsymbol{x}_{i}(\boldsymbol{x}_{0},\boldsymbol{\epsilon}),i)\right\|_{2}^{2}\right]+C
\end{gathered}
\end{equation}

where \(L_(t-1)\) is a representative term in the ELBO optimization objective for variational inference (see Eq.8 in \cite{ref1}), and \(C\) is a constant that does not depend on \(\theta\), \(\boldsymbol{\epsilon}{\sim}N(\boldsymbol{O},\boldsymbol{I}),~\boldsymbol{x}_i(\boldsymbol{x}_0, \boldsymbol{\epsilon})=\boldsymbol{x}_0+\sigma_i\boldsymbol{\epsilon} \). Therefore, we can parameterize \(\mu_\theta(x_i,i)\)

\begin{equation}
\mu_\theta(\boldsymbol{x}_i,i)=\boldsymbol{x}_i-({\sigma_i}^2-{\sigma_{i-1}}^2)\frac{\epsilon_\theta(\boldsymbol{x})}{\sigma_i}
\end{equation}

From Eq.\ref{eq:5} we convert the above equation from a noise prediction model to a data prediction model

\begin{equation}
\begin{aligned}
\mu_\theta({\boldsymbol{x}_i},i)&={\boldsymbol{x}_i}-({\sigma_i}^2-{\sigma_{i-1}}^2)\frac{\boldsymbol{x}_i-D_\theta(\boldsymbol{x})}{{\sigma_i}^2}\\&=\frac{{\sigma_{i-1}}^2}{{\sigma_i}^2}\boldsymbol{x}_i+\left(1-\frac{{\sigma_{i-1}}^2}{{\sigma_i}^2}\right)D_\theta(\boldsymbol{x})
\end{aligned}
\end{equation}

where \(D_\theta(\boldsymbol{x}_i )\) is the estimation of the denoising result of image \(\boldsymbol{x}_i\), in the simplest case \(D_\theta(\boldsymbol{x})\) = \(D_\theta(\boldsymbol{x}_i, i)\). Here we generalize it to \(\boldsymbol{x}_{j\in\{0,...,N\}}\), then \(D_\theta(\boldsymbol{x})=\boldsymbol{c}(i)\boldsymbol{D}_\theta(x_j)\) satisfies \(\boldsymbol{x}_{i}=\left\{\begin{matrix}{\boldsymbol{x}_{i},\boldsymbol{x}\leq i}\\{\hat{\boldsymbol{x}}_{i},\boldsymbol{x}>i}\\\end{matrix}\right.,\hat{\boldsymbol{x}}_{i+k}=\frac{\sigma_{i+k}}{\sigma_{i}}\boldsymbol{x}_{i}+\left(1-\frac{\sigma_{i+k}}{\sigma_{i}}\right)D_{\theta}(\boldsymbol{x}_{i}),k\in(0,1]\) and \(\forall i,\sum c_j(i)=1\)\par

As described in \cite{ref1}, we let \(\tau_{i}=\frac{\sigma_{i-1}}{\sigma_{i}}\sqrt{{\sigma_{i}}^{2}-{\sigma_{i-1}}^{2}}\). Through the ancestral sampling of \(\prod_{i=1}^Np(\boldsymbol{x}_{i-1}|\boldsymbol{x}_i)\), we obtain the following iteration rule

\begin{equation}
\boldsymbol{x}_{i+1}=\frac{\sigma_{i+1}^2}{\sigma_i^2}\boldsymbol{x}_i+\left(1-\frac{\sigma_{i+1}^2}{\sigma_i^2}\right)\boldsymbol{c}_(i)\boldsymbol{D}_\theta(\boldsymbol{x}_i)+\frac{\sigma_{i+1}}{\sigma_i}\sqrt{\sigma_i^2-\sigma_{i+1}^2}\boldsymbol{\epsilon}_i
\end{equation}

where \(x_N{\sim}\mathcal{N}(\boldsymbol{O},{\sigma_N}^2\boldsymbol{I})\mathrm{~.~}\boldsymbol{\epsilon}{\sim}\mathcal{N}(\boldsymbol{O},\boldsymbol{I})\)

\section{The Value of c for Common Samplers}
\label{app:c}

We focus on several ODE and SDE samplers proposed by Kerras et al. and Lu et al.. Considering comparisons in the same application scenarios, we refer to their implementations in the kdiffusion repository and use the same default values in the code implementation for some optional hyperparameters.

\begin{table}[htbp]
\centering
\begin{tabular}{ |l|c|c| } 
\hline
Sampler & \(j\) & \(c_j\) \\
\hline
\multirow{2}{6em}{Euler} & \(i\) & 1 \\ 
& else & 0 \\ 
\hline
\multirow{3}{6em}{Heun} & \(i\) & \(1-\frac{\sigma_i}{2\sigma_{i+1}}\) \\ [1.5ex]
& \(i+1\) & \(\frac{\sigma_i}{2\sigma_{i+1}}\)\\[1.5ex]
& else & 0 \\ 
\hline
\multirow{3}{6em}{DPM++ 2M} & \(i\) & \(\begin{aligned}1+\frac{\frac{\log\sigma_i-\log\sigma_{i-1}}{\log\sigma_{i+1}-\log\sigma_i}}{2\frac{\sigma_i-\sigma_{i+1}}{\sigma_i}}\end{aligned}\) \\ [2.5ex]
& \(i-1\) &\(\begin{aligned}-\frac{\frac{\log\sigma_i-\log\sigma_{i-1}}{\log\sigma_{i+1}-\log\sigma_i}}{2\frac{\sigma_{i-\sigma_{i+1}}}{\sigma_i}}\end{aligned}\)\\[2.5ex]
& else & 0 \\ 
\hline
\multirow{3}{6em}{DPM2} & \(i\) & \(1-\sqrt{\frac{\sigma_i}{\sigma_{i+1}}}\) \\ [1.5ex]
& \(i+\frac12\) & \(\sqrt{\frac{\sigma_i}{\sigma_{i+1}}}\)\\[1.5ex]
& else & 0 \\ 
\hline
\end{tabular}
\caption{c in ODE Samplers}
\end{table}

\begin{table}[htbp]
\centering
\begin{tabular}{ |l|c|c| } 
\hline
Sampler & \(j\) & \(c_j\) \\
\hline
\multirow{2}{6em}{Euler a} & \(i\) & 1 \\ 
& else & 0 \\ 
\hline
\multirow{3}{6em}{DPM++ SDE} & \(i\) & \(\frac{1}{2}\) \\ [1.5ex]
& \(i+1\) & \(\frac{1}{2}\)\\[1.5ex]
& else & 0 \\ 
\hline
\multirow{3}{6em}{DPM++ 2S a} & \(i+1\) & 1 \\[1.5ex]
& else & 0 \\ 
\hline
\multirow{3}{6em}{DPM++ 2M SDE} & \(i\) & \(1+\frac12\frac{\log\frac{\sigma_i}{\sigma_{i+1}}}{\log\frac{\sigma_{i-1}}{\sigma_i}}\) \\ [2.5ex]
& \(i-1\) & \(\-\frac12\frac{\log\frac{\sigma_i}{\sigma_{i+1}}}{\log\frac{\sigma_{i-1}}{\sigma_i}}\)\\[2.5ex]
& else & 0 \\ 
\hline
\multirow{3}{6em}{DPM2 a} & \(i\) & \(1-\frac{\sigma_i}{\sigma_{i+1}}\) \\ [1.5ex]
& \(i+1\) & \(\frac{\sigma_i}{\sigma_{i+1}}\)\\[1.5ex]
& else & 0 \\ 
\hline
\end{tabular}
\caption{c in SDE Samplers}
\end{table}

\section{Experimental Details}
In this section, we detail the specific environment and parameter configuration for the experimental part. All experiments were performed on an NVIDIA 3060.

\subsection{Configurations for Baselines}
We choose Euler method, Heun second-order sampler, DPM++ multistep sampler DPM++ 2M proposed in DPM and DPM2 second-order sampler proposed in DPM as the ODE baseline and Euler a, DPM2 a, DPM++ 2S a and DPM++ SDE as the SDE baseline.\par
We borrow hyperparameters from the Stable Diffusion model in the diffusers code repository, such as the discretization scheme or the initial noise scale. We use the DDIM and DDPM samplers implemented in the repository directly.

\subsection{Configurations for Sampler Scheduler}
We tested the following Sampler Scheduler settings separately for NFE=6N during our experiments:

\begin{center}
\begin{longtable}{|c|c|c|c|c|}
\hline
Unit1 Steps & Unit1 Sampler & Unit2 Steps & Unit2 Sampler & Type\\
\hline
\([1, N]\) & Heun & \([N+1, 5N+1]\) & Euler & ODE\\
\([1, 2N]\) & Heun & \([2N+1, 4N+1]\) & Euler & ODE\\
\([1, N]\) & Heun & \([N+1, 5N+1]\) & Euler a & ODE+SDE\\
\([1, 2N]\) & Heun & \([2N+1, 4N+1]\) & Euler a & ODE+SDE\\
\([1, N]\) & Heun & \([N+1, 5N+1]\) & DPM++ 2M & ODE\\
\([1, 2N]\) & Heun & \([2N+1, 4N+1]\) & DPM++ 2M & ODE\\
\([1, N]\) & Heun & \([N+1, 3N+1]\) & DPM2 & ODE\\
\([1, 2N]\) & Heun & \([2N+1, 3N+1]\) & DPM2 & ODE\\
\([1, N]\) & Heun & \([N+1, 3N+1]\) & DPM2 a & ODE+SDE\\
\([1, 2N]\) & Heun & \([2N+1, 3N+1]\) & DPM2 a & ODE+SDE\\
\([1, N]\) & Heun & \([N+1, 3N+1]\) & DPM++ 2S a & ODE+SDE\\
\([1, 2N]\) & Heun & \([2N+1, 3N+1]\) & DPM++ 2S a & ODE+SDE\\
\([1, N]\) & Heun & \([N+1, 3N+1]\) & DPM2++ SDE & ODE+SDE\\
\([1, 2N]\) & Heun & \([2N+1, 3N+1]\) & DPM++ SDE & ODE+SDE\\

\([1, N]\) & DPM2 & \([N+1, 5N+1]\) & Euler & ODE\\
\([1, 2N]\) & DPM2 & \([2N+1, 4N+1]\) & Euler & ODE\\
\([1, N]\) & DPM2 & \([N+1, 5N+1]\) & Euler a & ODE+SDE\\
\([1, 2N]\) & DPM2 & \([2N+1, 4N+1]\) & Euler a & ODE+SDE\\
\([1, N]\) & DPM2 & \([N+1, 5N+1]\) & DPM++ 2M & ODE\\
\([1, 2N]\) & DPM2 & \([2N+1, 4N+1]\) & DPM++ 2M & ODE\\
\([1, N]\) & DPM2 & \([N+1, 3N+1]\) & Heun & ODE\\
\([1, 2N]\) & DPM2 & \([2N+1, 3N+1]\) & Heun & ODE\\
\([1, N]\) & DPM2 & \([N+1, 3N+1]\) & DPM2 a & ODE+SDE\\
\([1, 2N]\) & DPM2 & \([2N+1, 3N+1]\) & DPM2 a & ODE+SDE\\
\([1, N]\) & DPM2 & \([N+1, 3N+1]\) & DPM++ 2S a & ODE+SDE\\
\([1, 2N]\) & DPM2 & \([2N+1, 3N+1]\) & DPM++ 2S a & ODE+SDE\\
\([1, N]\) & DPM2 & \([N+1, 3N+1]\) & DPM2++ SDE & ODE+SDE\\
\([1, 2N]\) & DPM2 & \([2N+1, 3N+1]\) & DPM++ SDE & ODE+SDE\\

\([1, N]\) & DPM2 a & \([N+1, 5N+1]\) & Euler & ODE+SDE\\
\([1, 2N]\) & DPM2 a & \([2N+1, 4N+1]\) & Euler & ODE+SDE\\
\([1, N]\) & DPM2 a & \([N+1, 5N+1]\) & Euler a & SDE\\
\([1, 2N]\) & DPM2 a & \([2N+1, 4N+1]\) & Euler a & SDE\\
\([1, N]\) & DPM2 a & \([N+1, 5N+1]\) & DPM++ 2M & ODE+SDE\\
\([1, 2N]\) & DPM2 a & \([2N+1, 4N+1]\) & DPM++ 2M & ODE+SDE\\
\([1, N]\) & DPM2 a & \([N+1, 3N+1]\) & Heun & ODE+SDE\\
\([1, 2N]\) & DPM2 a & \([2N+1, 3N+1]\) & Heun & ODE+SDE\\
\([1, N]\) & DPM2 a & \([N+1, 3N+1]\) & DPM2 & ODE+SDE\\
\([1, 2N]\) & DPM2 a & \([2N+1, 3N+1]\) & DPM2 & ODE+SDE\\
\([1, N]\) & DPM2 a & \([N+1, 3N+1]\) & DPM++ 2S a & SDE\\
\([1, 2N]\) & DPM2 a & \([2N+1, 3N+1]\) & DPM++ 2S a & SDE\\
\([1, N]\) & DPM2 a & \([N+1, 3N+1]\) & DPM2++ SDE & SDE\\
\([1, 2N]\) & DPM2 a & \([2N+1, 3N+1]\) & DPM++ SDE & SDE\\

\([1, N]\) & DPM++ 2S a & \([N+1, 5N+1]\) & Euler & ODE+SDE\\
\([1, 2N]\) & DPM++ 2S a & \([2N+1, 4N+1]\) & Euler & ODE+SDE\\
\([1, N]\) & DPM++ 2S a & \([N+1, 5N+1]\) & Euler a & SDE\\
\([1, 2N]\) & DPM++ 2S a & \([2N+1, 4N+1]\) & Euler a & SDE\\
\([1, N]\) & DPM++ 2S a & \([N+1, 5N+1]\) & DPM++ 2M & ODE+SDE\\
\([1, 2N]\) & DPM++ 2S a & \([2N+1, 4N+1]\) & DPM++ 2M & ODE+SDE\\
\([1, N]\) & DPM++ 2S a & \([N+1, 3N+1]\) & Heun & ODE+SDE\\
\([1, 2N]\) & DPM++ 2S a & \([2N+1, 3N+1]\) & Heun & ODE+SDE\\
\([1, N]\) & DPM++ 2S a & \([N+1, 3N+1]\) & DPM2 & ODE+SDE\\
\([1, 2N]\) & DPM++ 2S a & \([2N+1, 3N+1]\) & DPM2 & ODE+SDE\\
\([1, N]\) & DPM++ 2S a & \([N+1, 3N+1]\) & DPM2 a & SDE\\
\([1, 2N]\) & DPM++ 2S a & \([2N+1, 3N+1]\) & DPM2 a & SDE\\
\([1, N]\) & DPM++ 2S a & \([N+1, 3N+1]\) & DPM2++ SDE & SDE\\
\([1, 2N]\) & DPM++ 2S a & \([2N+1, 3N+1]\) & DPM++ SDE & SDE\\

\([1, N]\) & DPM++ SDE & \([N+1, 5N+1]\) & Euler & ODE+SDE\\
\([1, 2N]\) & DPM++ SDE & \([2N+1, 4N+1]\) & Euler & ODE+SDE\\
\([1, N]\) & DPM++ SDE & \([N+1, 5N+1]\) & Euler a & SDE\\
\([1, 2N]\) & DPM++ SDE & \([2N+1, 4N+1]\) & Euler a & SDE\\
\([1, N]\) & DPM++ SDE & \([N+1, 5N+1]\) & DPM++ 2M & ODE+SDE\\
\([1, 2N]\) & DPM++ SDE & \([2N+1, 4N+1]\) & DPM++ 2M & ODE+SDE\\
\([1, N]\) & DPM++ SDE & \([N+1, 3N+1]\) & Heun & ODE+SDE\\
\([1, 2N]\) & DPM++ SDE & \([2N+1, 3N+1]\) & Heun & ODE+SDE\\
\([1, N]\) & DPM++ SDE & \([N+1, 3N+1]\) & DPM2 & ODE+SDE\\
\([1, 2N]\) & DPM++ SDE & \([2N+1, 3N+1]\) & DPM2 & ODE+SDE\\
\([1, N]\) & DPM++ SDE & \([N+1, 3N+1]\) & DPM2 a & SDE\\
\([1, 2N]\) & DPM++ SDE & \([2N+1, 3N+1]\) & DPM2 a & SDE\\
\([1, N]\) & DPM++ SDE & \([N+1, 3N+1]\) & DPM2++ 2S a & SDE\\
\([1, 2N]\) & DPM++ SDE & \([2N+1, 3N+1]\) & DPM++ 2S a & SDE\\
\hline
\caption{Sampler Scheduler settings}
\end{longtable}
\end{center}

Based on experience and experimental results (see Appendix \ref{app:e} for details), we believe that DPM2 a[N] DPM2[2N], DPM2 a[N] DPM++2M[4N], DPM++ 2S a[N] DPM2[2N], DPM++ 2S a[N] DPM++2M[4N], DPM++ SDE[N] DPM++2M[4N] and DPM ++ SDE[N] DPM2[2N] are the better performing configurations. Our composite Sampler Scheduler score is mainly obtained in these configurations.\par
For simplicity purposes, we mainly validate the strategy of scheduling two different samplers by dividing the sampling step into two parts. More splits and more complex scheduling strategies are subject to further research.

\subsection{Pre-trained Models}
For the CIFAR-10 dataset, we use the pre-trained VP and EDM models from the EDM repository.

\section{More Results}
\label{app:e}
In this section, we provide more detailed results, both quantitative and qualitative.

\subsection{Experiments on Standard Benchmarks}

\begin{center}
\begin{longtable}{|l|c|c|c|c|c|}
\hline
SamplerSchedule Config & N=1 & N=2 & N=3 & N=4 & N=5\\
\hline
Euler[6N] & 36.9753 & 11.9055 & 7.02887 & 5.15021 & 4.20923 \\
Heun[3N] & 122.58 & 36.9753 & 18.2181 & 11.9018 & 8.8082 \\
DPM++2M[6N] & 17.1086 & 3.73001 & 2.45117 & 2.16493 & 2.06226\\
DPM2[3N] & 4.35493 & 4.36003 & 1.93805 & 1.97308 & 1.94304\\
Euler a[6N] & 66.7995 & 37.0076 & 25.0982 & 18.6362 & 14.5688\\
DPM2 a[3N] & 343.968 & 70.4262 & 7.35205 & 3.14369 & 3.58586\\
DPM++ 2S a[3N] & 116.694 & 36.1001 & 19.9291 & 13.3843 & 9.74262\\
DPM++ SDE[3N] & 126.194 & 50.6659 & 31.9771 & 23.1409 & 17.807\\
Heun[N] Euler[4N] & 46.5607 & 15.3784 & 8.75686 & 5.89091 & 4.76504\\
Heun[2N] Euler[2N] & 36.688 & 12.3641 & 8.8742 & 6.7805 & 5.49571\\
Heun[N] Euler a[4N] & 75.8231 & 43.1475 & 30.0876 & 21.5624 & 16.9991\\
Heun[2N] Euler a[2N] & 37.0844 & 31.3069 & 28.629 & 24.4145 & 20.2743\\
Heun[N] DPM++2M[4N] & 26.7362 & 5.0428 & 2.82565 & 2.26334 & 2.11961\\
Heun[2N] DPM++2M[2N] & 36.688 & 6.72377 & 3.64267 & 2.54556 & 2.26837\\
Heun[N] DPM2[2N] & 269.058 & 5.13605 & 2.27613 & 1.99763 & 1.96439\\
Heun[2N] DPM2[N] & 305.151 & 13.6177 & 2.78049 & 1.91444 & 1.86846\\
Heun[N] DPM2 a[2N] & 354.822 & 70.7907 & 7.06029 & 3.23029 & 3.80714\\
Heun[2N] DPM2 a[N] & 305.151 & 112.857 & 12.5293 & 2.17304 & 2.24146\\
Heun[N] DPM++ 2S a[2N] & 84.7442 & 33.8302 & 19.9419 & 12.0496 & 8.93943\\
Heun [2N] DPM++ 2S a [N] & 305.151 & 10.6309 & 5.26631 & 7.10264 & 5.8395\\
Heun[N] DPM++ SDE[2N] & 84.7722 & 46.4282 & 30.9697 & 21.1332 & 16.63\\
Heun[2N] DPM++ SDE[N] & 305.151 & 10.8013 & 6.34447 & 10.7785 & 9.47347\\
DPM2[N] Euler[4N] & 47.9796 & 15.452 & 8.7663 & 5.89513 & 4.76648\\
DPM2[2N] Euler[2N] & 23.7705 & 14.2842 & 9.46983 & 6.92793 & 5.57545\\
DPM2[N] Euler a[4N] & 76.7837 & 42.9862 & 30.1833 & 21.3235 & 17.0488\\
DPM2[2N] Euler a[2N] & 24.4076 & 34.0571 & 29.6189 & 24.7088 & 20.4931\\
dpm2[n] dpm++2m[4n] & 27.3529 & 5.00791 & 2.81582 & 2.25904 & 2.11805\\
dpm2[2n] dpm++2m[2n] & 23.7705 & 7.34738 & 3.72921 & 2.56704 & 2.28787\\
DPM2[N] Heun[2N] & 76.995 & 34.2494 & 17.5499 & 10.491 & 8.01974\\
DPM2[2N] Heun[N] & 149.046 & 6.77367 & 5.71256 & 6.02181 & 5.06405\\
DPM2[N] DPM2 a[2N] & 312.688 & 71.0221 & 7.22097 & 3.14962 & 3.76813\\
DPM2[2N] DPM2 a[N] & 149.046 & 95.8176 & 11.0185 & 2.10185 & 2.30299\\
DPM2[N] DPM++ 2S a[2N] & 77.3944 & 34.5697 & 19.9083 & 12.092 & 8.96309\\
DPM2 [2N] DPM++ 2S a [N] & 149.046 & 6.82455 & 6.3872 & 7.71925 & 6.19253\\
dpm2[n] dpm++ sde[2n] & 77.5796 & 47.6591 & 31.5243 & 21.0063 & 16.6248\\
dpm2[2n] dpm++ sde[n] & 149.046 & 7.06167 & 7.78363 & 11.6859 & 10.0291\\
DPM2 a[N] Euler[4N] & 48.3822 & 15.4605 & 8.76381 & 5.98218 & 4.82529\\
DPM2 a[2N] Euler[2N] & 70.8603 & 18.0002 & 10.794 & 7.46902 & 6.15688\\
DPM2 a[N] Euler a[4N] & 76.6334 & 43.3081 & 30.0343 & 21.4837 & 17.0927\\
DPM2 a[2N] Euler a[2N] & 71.0415 & 37.8954 & 31.1961 & 25.0199 & 20.6227\\
DPM2 a[N] DPM++2M[4N] & 27.1482 & 5.02655 & 2.75532 & 2.26565 & 2.07016\\
DPM2 a[2N] DPM++2M[2N] & 71.0175 & 9.76274 & 4.43236 & 2.82717 & 2.46283\\
DPM2 a[N] Heun[2N] & 111.757 & 36.2729 & 18.1277 & 10.4727 & 8.24126\\
DPM2 a[2N] Heun[N] & 343.898 & 27.4138 & 10.2951 & 8.98419 & 7.56196\\
DPM2 a[N] DPM2[2N] & 285.082 & 4.484 & 2.08661 & 1.89887 & 1.84535\\
DPM2 a[2N] DPM2[N] & 343.861 & 25.8825 & 3.95346 & 3.27759 & 3.3147\\
DPM2 a[N] DPM++ 2S a[2N] & 111.558 & 36.3644 & 20.0079 & 12.0186 & 8.87151\\
DPM2 a[2N] DPM++ 2S a [N] & 343.992 & 27.6426 & 10.9206 & 10.6151 & 8.61023\\
DPM2 a[N] DPM++ SDE[2N] & 111.873 & 49.6038 & 31.7008 & 20.9955 & 16.5121\\
DPM2 a[2N] DPM++ SDE[N] & 343.846 & 27.7593 & 12.6424 & 15.1028 & 12.7868\\
DPM++ 2S a[N] Euler[4N] & 48.7258 & 15.6103 & 8.92208 & 5.95825 & 4.78529\\
DPM++ 2S a[2N] Euler[2N] & 63.8916 & 21.0826 & 11.2175 & 7.2695 & 5.75928\\
DPM++ 2S a[N] Euler a[4N] & 76.6333 & 43.3433 & 29.9405 & 21.466 & 17.1479\\
DPM++ 2S a[2N] Euler a[2N] & 64.8939 & 42.0342 & 31.6934 & 25.1514 & 20.2093\\
DPM++ 2S a[N] DPM++2M[4N] & 27.5243 & 4.92915 & 2.76419 & 2.22723 & 2.10222\\
DPM++ 2S a[2N] DPM++2M[2N] & 63.5326 & 11.4574 & 4.63398 & 2.68966 & 2.36079\\
DPM++ 2S a[N] Heun[2N] & 115.345 & 36.404 & 18.0675 & 10.6672 & 8.04489\\
DPM++ 2S a[2N] Heun[N] & 116.862 & 35.912 & 18.9764 & 10.4107 & 7.88842\\
DPM++ 2S a[N] DPM2[2N] & 110.546 & 5.64848 & 2.09696 & 1.91134 & 1.91003\\
DPM++ 2S a[2N] DPM2[N] & 116.585 & 30.982 & 10.6693 & 4.42001 & 3.57323\\
DPM++ 2S a[N] DPM2 a[2N] & 298.559 & 74.7157 & 7.75672 & 3.12054 & 3.75969\\
DPM++ 2S a[2N] DPM2 a[N] & 116.82 & 143.663 & 19.091 & 4.1801 & 3.98827\\
DPM++ 2S a[N] DPM++ SDE[2N] & 117.547 & 49.7391 & 31.4504 & 21.0851 & 16.4997\\
DPM++ 2S a[2N] DPM++ SDE[N] & 116.9 & 36.7249 & 22.2288 & 16.57 & 13.4113\\
DPM++ SDE[N] Euler[4N] & 48.8996 & 15.6749 & 8.74443 & 5.90989 & 4.75383\\
DPM++ SDE[2N] Euler[2N] & 78.1132 & 25.6867 & 13.3436 & 7.87875 & 6.20821\\
DPM++ SDE[N] Euler a[4N] & 77.7119 & 43.0762 & 29.9269 & 21.4137 & 17.0894\\
DPM++ SDE[2N] Euler a[2N] & 79.4571 & 46.8062 & 34.0847 & 25.6235 & 20.6207\\
dpm++ sde[n] dpm++2m[4n] & 27.7884 & 4.90291 & 2.77185 & 2.22644 & 2.09108\\
dpm++ sde[2n] dpm++2m[2n] & 78.2014 & 14.1965 & 5.96189 & 3.16279 & 2.71877\\
DPM++ SDE[N] Heun[2N] & 123.631 & 37.4668 & 18.4202 & 10.7238 & 8.16934\\
DPM++ SDE[2N] Heun[N] & 125.308	& 50.0675 & 28.1936 & 14.725 & 11.081\\
DPM++ SDE[N] DPM2[2N] & 109.103 & 6.57066 & 2.3558 & 1.9752 & 1.93999\\
DPM++ SDE[2N] DPM2[N] & 125.63 & 44.8777 & 18.0953 & 7.48713 & 6.06719\\
DPM++ SDE[N] DPM2 a[2N] & 297.214 & 77.0681 & 8.34158 & 3.26922 & 3.7997\\
DPM++ SDE[2N] DPM2 a[N] & 125.892 & 152.034 & 27.532 & 6.72327 & 6.1664\\
DPM++ SDE[N] DPM++ 2S a[2N] & 125.297 & 37.2987 & 20.1741 & 11.9307 & 8.75621\\
DPM++ SDE [2N] DPM++ 2S a [N] & 126.237 & 50.161 & 29.574 & 15.7865 & 11.6821\\
\hline
\caption{FID result}
\end{longtable}
\end{center}

\subsection{Experiments on Large-scale Text-to-image Model}

\begin{table}[htbp]
\centering
\begin{tabular}{|l|c|c|c|c|}
\hline
Sampler & w=2 & w=3 & w=5 & w=8\\
\hline
DDIM[45] & 0.2901 & 0.3056 & 0.3152 & 0.3206\\
DDIM[90] & 0.2918 & 0.3056 & 0.3157 & 0.32\\
Heun[46] & 0.286 & 0.3022 & 0.3135 & 0.3185\\
Heun[91] & 0.2869 & 0.3025 & 0.3135 & 0.3185\\
DDPM[90] & 0.3012 & 0.3128 & 0.3196 & 0.3222\\
DDPM[180] & 0.3005 & 0.3131 & 0.3199 & 0.3212\\
Sampler Scheduler[91] & 0.2979 & 0.3062 & 0.318 & 0.3224\\
\hline
\end{tabular}
\caption{CLIP Score}
\end{table}

\begin{table}[!h]
\centering
\begin{tabular}{|l|c|c|c|c|}
\hline
Sampler & w=2 & w=3 & w=5 & w=8\\
\hline
DDIM[45] & 5.13 & 5.21 & 5.3 & 5.37\\
DDIM[90] & 5.15 & 5.22 & 5.31 & 5.36\\
Heun[46] & 5.14 & 5.18 & 5.12 & 5.33\\
Heun[91] & 5.14 & 5.2 & 5.28 & 5.33\\
DDPM[90] & 5.2 & 5.29 & 5.34 & 5.39\\
DDPM[180] & 5.19 & 5.28 & 5.33 & 5.37\\
Sampler Scheduler[91] & 5.19 & 5.27 & 5.32 & 5.4\\
\hline
\end{tabular}
\caption{Aesthetic Score}
\end{table}

\section{Extended Generated Images}
In this section, we provide extensions to generate images via Sampler Scheduler (Seniorious) and other mainstream samplers on the text-to-image model Stable Diffusion v1.5 and the widely used model Anything-v5-PrtRE by the community.

\begin{figure}[htbp]
\centering
{\includegraphics[width=.8\columnwidth]{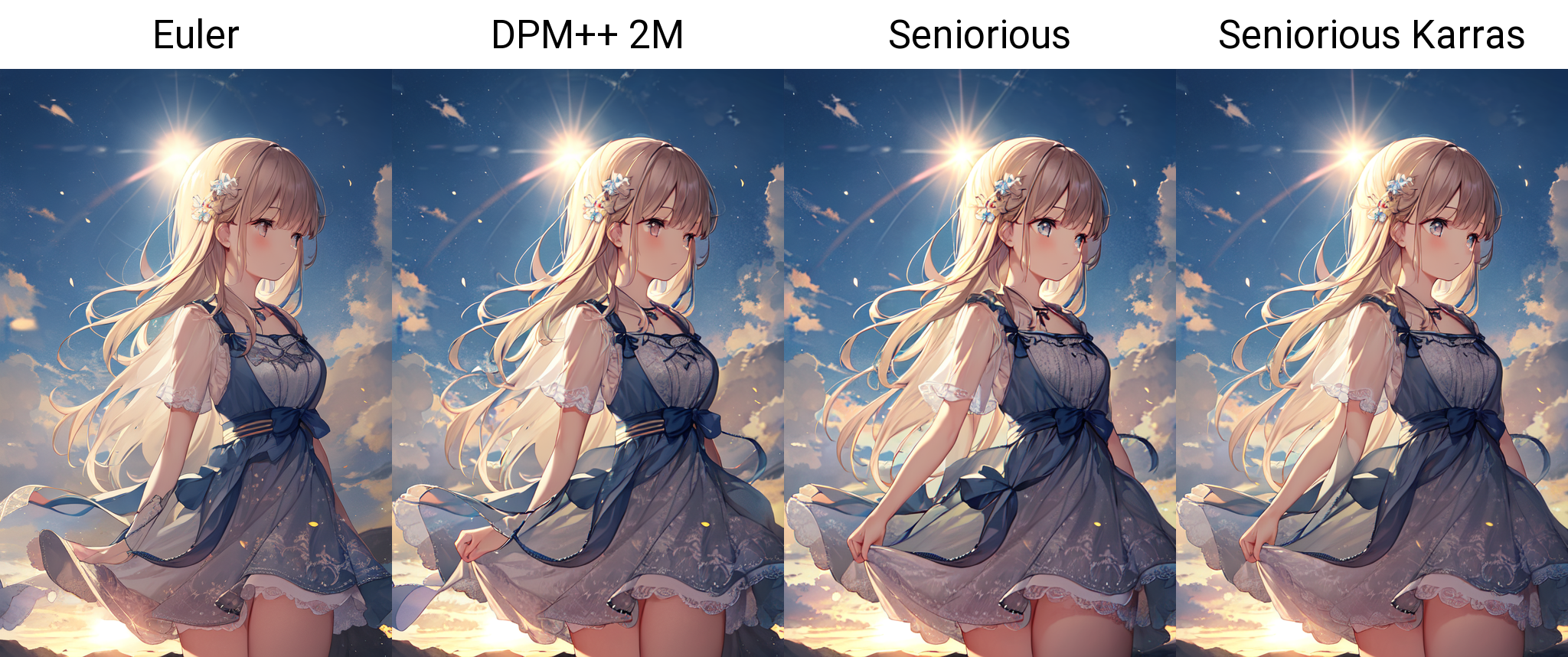}}
\label{fig:5}
\caption{Euler, DPM++ 2M and ODE Sampler Scheduler (Seniourious)}
\end{figure}

\begin{figure}[htbp]
\centering
{\includegraphics[width=.8\columnwidth]{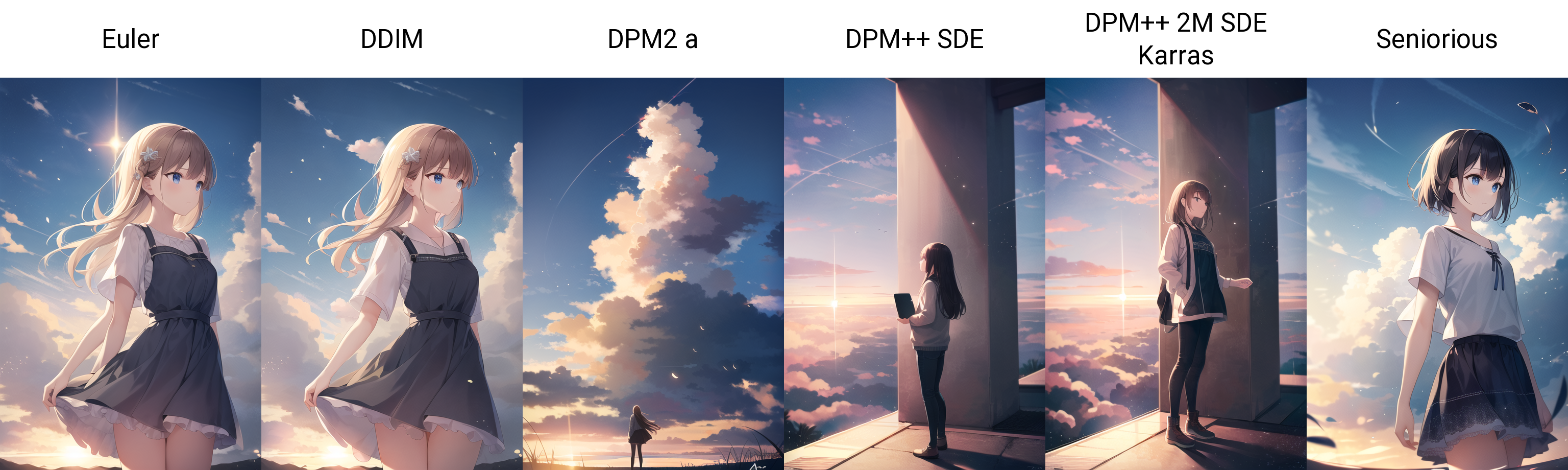}}
\label{fig:6}
\caption{Euler, DDIM, DPM2 a, DPM++ SDE, DPM++ 2M SDE and Mixed Sampler Scheduler (Seniourious)}
\end{figure}

\end{document}